# Dynamic Modelling and Adaptive Traction Control for Mobile Robots

A. Albagul[1] & Wahyudi[2]
[1] IIUM, Mechatronics Engineering Department, albagul@iiu.edu.my
[2] IIUM, Mechatronics Engineering Department, wahyudi@iiu.edu.my

*Abstract: Mobile robots have received a great deal of research in recent years. A significant amount of research has been published in many aspects related to mobile robots. Most of the research is devoted to design and develop some control techniques for robot motion and path planning. A large number of researchers have used kinematic models to develop motion control strategy for mobile robots. Their argument and assumption that these models are valid if the robot has low speed, low acceleration and light load. However, dynamic modelling of mobile robots is very important as they are designed to travel at higher speed and perform heavy duty work. This paper presents and discusses a new approach to develop a dynamic model and control strategy for wheeled mobile robot which I modelled as a rigid body that roles on two wheels and a castor. The motion control strategy consists of two levels. The first level is dealing with the dynamic of the system and denoted as 'Low' level controller. The second level is developed to take care of path planning and trajectory generation.*
***Keywords***: *mobile robots, path planning, modelling, dynamic, adaptive control.*

## 1. Introduction

Mobile robots have been used in many application such as moving material between work stations. They can also be found in many areas such as industrial, medical, environmental and even domestic machines. Research on mobile robots has mounted and attracted so much attention in recent years since they are increasingly used in wide range of applications [1, 2, 3, 4]. At the beginning, most of research have been directed to the use of kinematic models of the mobile robots to achieve and accomplished the motion control [3, 4, 5]. Later on, the research has taken another approach and has focused on robots with additional sensory system to develop autonomous guidance path planning systems [5]. This direction has led to produce sophisticated sensory systems that can learn about the operating environment and hence evaluating path constraints to the path planning objective itself [6]. However, some research has also addressed some topics related to dynamic characteristics of the motion which are essential to path tracking objective. Shiller [7] has studied the problem of computing suitable trajectories in the face of varying point and in any direction. In fact, this model is not restricted to wheeled robot. It can be used to obtain general free body motion dynamics. For the purpose of this work, the motive forces driving the robot are exerted by two independent DC motors each driving on of the rear wheels. This configuration is commonly used in terrain topography and under road holding constraints. The problems of road handling and traction become very important when the robot is subjected to dynamic variations. These variations include changes in robot inertia and centre of gravity which caused by the variable carrying load. The changes in the terrain topography, texture or in wheel properties due to wear, contamination or deformation play a major role in the robot motion. These variations can easily affect the traction properties and hence the robot movement and may cause slippage to occur. Therefore, it is very important for the robot to be able to avoid slippage and handle is consequences. This requires some learning mechanism which will try to adapt the trajectory planning strategy to cope with any condition. This paper presents the work that has been done to explore the issue of dynamic modelling and motion control under varying loading conditions. This leads to the development of a dynamic model for a three wheeled mobile robot structure [1, 2]. The model includes the capability for representing any shape of cart with variable density to accommodate any changes in the robot structure. The motion dynamic for the robot can be achieved by applying a number of forces acting at any autonomous vehicles and is called differential drive mechanism. The simulation model of the robot dynamics is implemented in the Matlab computational environment. The exploration of the inertia variation and traction conditions while using standard motor control algorithms for given motion trajectories will be presented and discussed. The over all task of motion control eventually translates into deriving the required input to the DC motors. This can be divided into kinematics path trajectory planning and translation into the dynamic motor rotation trajectories. In order to execute accurate paths, the motor motion trajectories must consider both constraints do to the motor and load dynamics and those due to chrematistics. Elements of feed forward (pre-planning) and feedback are required to optimize the trajectories. Cross coupling effects between the two drive motors can also be compensated for and this provides for a further degree of accuracy in the path following objective In the face of varying conditions, autonomous adaptation requires that some measure of performance is made available. For the vehicle this will be in the form of position, orientation and wheel speed measurements. These provide feedback on dynamic path tracking accuracy and also on slippage. This feedback will enable both 'low' and 'high' level adaption of the vehicle motion in order to suit new conditions thus enabling the motion to be optimized continuously without reduction in efficiency and/or loss of stability.

**2. Robot Model**

The considered robot is modelled as a three dimensional rigid body which consists of a number of small particles connected together. The particles have a cubic shape with uniform density. This is to ease mathematical analysis which will be carried out on the body without reducing the generality of the model since the cubes can be made arbitrarily small and numerous. The shape and dimensional definition of each particle is shown in Fig. 1. By using the general model, the mass, inertia and centre of gravity of each particle can be calculated. Then the overall mass, inertia and centre of gravity of the entire body of the robot can be obtained. In Fig. 1, the axes of reference frame of the entire body are denoted as $X_0$, $Y_0$, and $Z_0$, while , $Y_n$, $Z_n$ are the axes of the reference frame for particle n. The original coordinates of each particle, n, are denoted as $X_{n0}$, $Y_{n0}$, $Z_{n0}$, while $X_{nmax}$, $Y_{nmax}$, $Z_{nmax}$ are the maximum dimensions of the particle relative to the origin of the particle reference frame. The centre of gravity of the particle to its reference frame is $X_{ncg}$, $Y_{ncg}$ and $Z_{ncg}$, while the center of gravity of the entire body is $X_{cg}$, $Y_{cg}$, $Z_{cg}$. Fig. 2 shows a 3D model of the robot created by Matlab tool.

*2.1. Kinematic Model*
Most of kinematic models of mobile robots assume that no tire slippage occurs, so the inputs to the system are right and left wheel angular velocities $\omega_r$ and $\omega_l$ respectively. Then the motion of the robot can be described by the simple kinematics of rigid bodies. In order to determine the robot motion, it is so important to define the position and orientation the robot as the location and orientation of the centre of gravity, p($X_0$, $Y_0$, $\theta_0$), relative to the fixed world frame. With the assumption of no lateral or longitudinal tire slip, the linear velocities of the right and left wheels can be expressed as:

$$v_r = R_t \omega_r \hat{i} \quad (1)$$

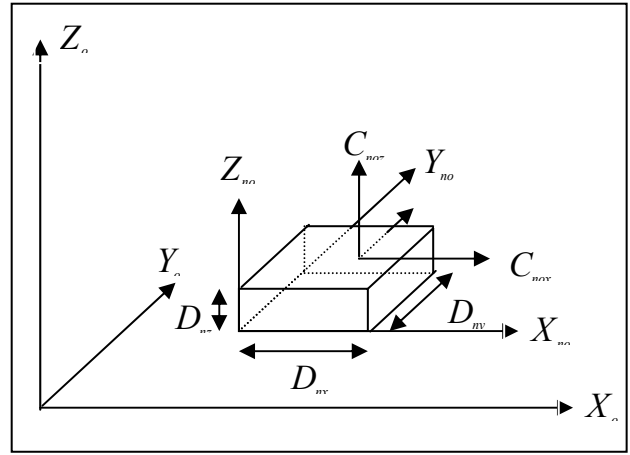

Fig. 1. Particle dimensional definition

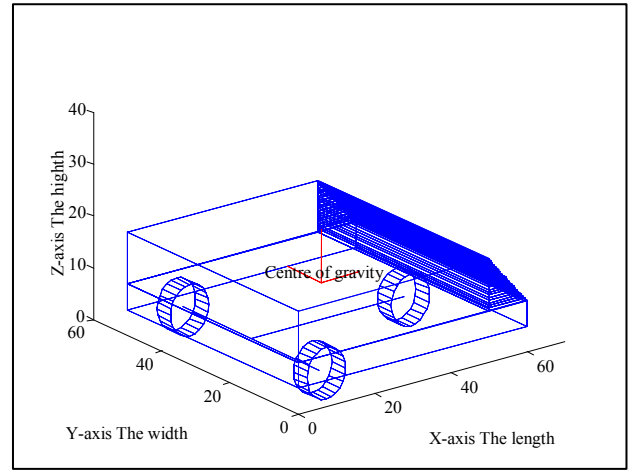

Fig. 2. A 3D model of the robot created by the Matlab

$$v_l = R_t \omega_l \hat{i} \quad (2)$$

The equations of motion of the rigid cart with respect to each wheel are:

$$V_R = V_g + r\hat{k}(-L_r\hat{i} - \frac{T_r}{2}\hat{j})$$
$$= u\hat{i} + v\hat{j} + r\frac{T_r}{2}\hat{i} - rL_r\hat{j} \quad (3)$$

$$V_L = V_g + r\hat{k}(-L_r\hat{i} + \frac{T_r}{2}\hat{j})$$
$$= u\hat{i} + v\hat{j} - r\frac{T_r}{2}\hat{i} - rL_r\hat{j} \quad (4)$$

where $V_g$ is the velocity vector of the centre of gravity, p($X_0$, $Y_0$, $\theta_0$), $L_R$ is the distance from the rear axle to the centre of gravity, $T_r$ id the distance between the two driving wheels, $\upsilon$, $v$ and $r$ are the forward, the lateral and the yaw velocities respectively. Using equations 1 -

4, the full kinematic model for the robot can be obtained as:

$$u = (\omega_r + \omega_l)\frac{R_t}{2} \quad (5)$$

$$r = (\omega_l - \omega_r)\frac{R_t}{T_r} \quad (6)$$

$$v = (\omega_l - \omega_r)\frac{L_R R_t}{T_r} \quad (7)$$

Then the heading, velocities and position of the robot in the world coordinate system can be obtained.

$$\vartheta_0 = \int r\, dt \quad (8)$$

$$V_X = u\cos\vartheta_0 - v\sin\vartheta_0 \quad (9)$$

$$V_Y = u\cos\vartheta_0 + v\sin\vartheta_0 \quad (10)$$

$$X_0 = \int V_X\, dt \quad (11)$$

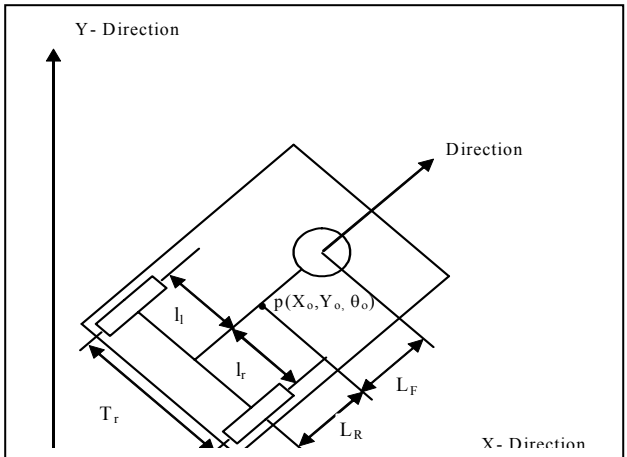

Fig. 3 Robot geometry and coordination

$$Y_0 = \int V_Y\, dt \quad (12)$$

where, $V_X$ and $V_Y$ are the velocity components of the vehicle relative to the world frame, $X_0$, $Y_0$ and $\theta_0$ are the position and heading of the robot in the world frame.

*2.2. Dynamic Model*

In order to obtain the dynamic model of the robot, forces must be applied, analyzed and moments must be taken about some point on the robot, in this case the centre of gravity. Since the robot is a three-degree of freedom 3DOF, which allows only movement in the longitudinal and lateral directions along with angular displacement, the equation of force and moment can be expressed as:

$$\sum F_X = m(\dot{u} - vr) \quad (13)$$

$$\sum F_Y = m(\dot{v} - ur) \quad (14)$$

$$\sum M_Z = I_Z \dot{r} \quad (15)$$

The forces acting on the robot are those forces exerted by the right and left driving wheels. These forces are proportional to the applied torque minus the amount of torque required to accelerate the wheels, gearbox and rotor. The applied torque is divided into two ways; linear torque to accelerate the robot and angular torque to accelerate the wheels, gearbox and rotor.

$$T_{app} = T_{lin} + T_{ang} \quad (16)$$

where $T_{app}$, $T_{lin}$ and $T_{ang}$ are the applied, the linear and the angular torques respectively.

The linear torque is converted into a longitudinal force at the tire/ground interface and expressed by:

$$F_X = \frac{T_{lin}}{R_t} \quad (17)$$

and the angular torque is:

$$T_{ang} = I_Z \dot{\omega} = I_Z \frac{\dot{u}}{R_t} \quad (18)$$

$$F_X = \frac{T_{app} - T_{ang}}{R_t} = \frac{R_t T_{app} - I_Z \dot{u}}{R_t^2} \quad (19)$$

Using equation 19 the distinction can be made to obtain the forces exerted be the right and left driving wheel as follows:

$$F_{Xr} = \frac{R_t T_{appr} - I_Z \dot{u}_r}{R_t^2} \quad (20)$$

$$F_{Xl} = \frac{R_t T_{appl} - I_Z \dot{u}_l}{R_t^2} \quad (21)$$

The dynamic equations describing the motion of the robot in terms of accelerations are:

$$\dot{V}_X = \frac{F_{Xr} + F_{Xl}}{m} + V_Y \omega \quad (22)$$

$$\dot{V}_Y = \frac{F_{Yr} + F_{Yl}}{m} - V_X \omega \quad (23)$$

$$\dot{r} = \frac{L_r F_{Xr} - L_l F_{Xl} - L_R (F_{Yr} + F_{Yl})}{I_Z} \quad (24)$$

The longitudinal, lateral and yaw velocities are simply the integral of their accelerations. Thus, the potion and heading of the robot can be obtained in the same manner as in the previous section.

**3. Path Planning**

Path planning is the subject that deals with the displacement of the robot in a prior known environment. It plays a major role in building an effective sophisticated mobile robot. Path planning as well as trajectory generation are required prior to the movement of the robot. The robot is desired to move from a starting position to a goal point in the workspace. A point on the robot is specified and designated to follow the required path. There are many methods to generate paths in terms of smoothness and curvature as well as continuity. Some of these methods are complicated and time consuming but produce very smooth paths. In this paper an effective method for path and trajectory generation is adopted. This method consists of straight lines joined by circular arc segments with a specified radius and turning angle. These circular segments are to avoid stoppage and provide continuity for the robot. This method is based on some parameters that should be known to define the path. These parameters are:
1- The start and end coordinates of each straight line section, or alternatively the length and heading of the section.
2- The start of each circular arc segment, its radius and turning angle, which corresponds to the change of orientation.

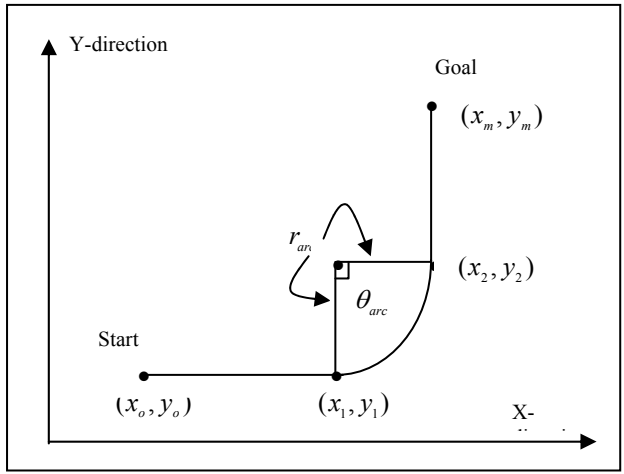

Fig. 4 Example of path definition

Fig. 4 shows a simple path that consists of two straight lines joined by an arc with a radius $r_{arc}$ and a turning angle, $\theta_{arc}$, of 90°.

*3.1. Velocity trajectory generation*
In this section, the generation of a velocity trajectory profile for a given continuous geometric path is treated as a function of time. As mentioned in the previous section that the path is divided into a number of sections connected together in series, each section can be set to have its own target of maximum velocity, called the section velocity. Also each section has been divided into three regions for acceleration, constant velocity and deceleration with respective lengths, $d_A$, $d_C$ and $d_D$ as well as time duration $t_A$, $t_C$ and $t_D$ plus angles and radius for the circular arc sections. The basis of the velocity trajectory is to accelerate in each section to reach the specified maximum velocity of that section if possible within the section distance. If the maximum velocity achieved for the section the robot must be able to decelerate to next maximum velocity of the next section to maintain the continuity of the path tracking. The intelligent of the algorithm is that it can work in no more than two operations. The first operation is to accelerate the robot and the second one is to decelerate it. The first operation will check if the maximum velocity of the section can be achieved, with the given acceleration, from the start velocity within the length of the section. If the maximum velocity can be reached then the acceleration distance and time are computed and the remaining distance of the section is driven at the maximum velocity. If the maximum velocity is not reachable, the robot will then either accelerate throughout the whole section if it is followed by another one or just accelerate and decelerate if it is one section. The exit velocity of a section must automatically be the start velocity of the next section. The second operation deals with calculating the distances and times for acceleration and deceleration from the start, section or the maximum velocity to the exit velocity of each section. If the section length is insufficient to decelerate to the exit velocity then the start velocity must be changed to another new velocity that the section can be completed with the desired exit velocity. It is required that the maximum velocity of the first section must be as low as possible for the robot to be able to decelerate normally to the start velocity of the next section if it is lower. A high maximum start velocity could only occur if the trajectory generation algorithm is used on-line.

*3.2. Explanation of the algorithm*
In the first step, each section is set to have a start velocity equal the previous section exit velocity for the sake of continuity and may be to avoid slippage if the robot is known to operate in some slippery sections or if it is executing a tight turn. If the start velocity is too high to decelerate to the exit velocity then it must be set to a lower velocity. The second step is to check if the length of the section large enough to allow the robot to reach for the maximum velocity if so then the time and distance, necessary for acceleration from the start velocity to the maximum velocity of the current section, are calculated as well as the time to drive the remaining distance with the constant maximum velocity. Otherwise the resultant velocity of constant acceleration is the end velocity of the section. First and second steps together ensure that there is no demand for velocity increase greater than the robot actuators can deliver with fixed acceleration. In the third step, the end velocity of the current section is set to equal to the start velocity of the next section. This is to maintain the smoothness and continuity of the motion and also that the deceleration in each section starts in time to complete the section and reach the start velocity of the next section. Forth step is dealing with calculating the time and distance needed for the robot to decelerate from the highest velocity reached during the execution of the section to its end velocity. The highest velocity is the maximum velocity if the section is long enough. Otherwise it is the peak velocity if the section contains only acceleration and deceleration parts.

**4. Motion Control**

The task of the controller is to achieve various goals and desired features for the robot motion. It is also designed to execute the planned sequences of motions correctly in the presence of any error. The control strategies in this work are based on controlling the traction forces causing the motion as well as maintaining the tracking of the desired path without any slippage or deviation. Traction motion control has some desired feature such as:
1- Maintaining the fastest possible acceleration and deceleration
2- Maintaining the desired path following accuracy
3- Maintaining the robot stability during the manoeuvres
4- Preventing the robot from slipping or sliding

The block diagram of the mobile robot control system is shown in Fig. 5. The controller tasks consist of two parts. The first part is the 'Low level' control strategy which deals with the dynamic changes in the robot. The second one is the 'High level' control strategy which deals with the changes in the environment in which the robot operates. The combined Low and High level motion control for the robot is depicted in Fig. 6.

*4.1. Low level control strategy*
In this strategy, an adaptive motion control algorithm based on the Pole placement self-tuning adaptive controller is considered. The controller has been designed with a PID structure to estimate the changes in the system dynamic parameters.

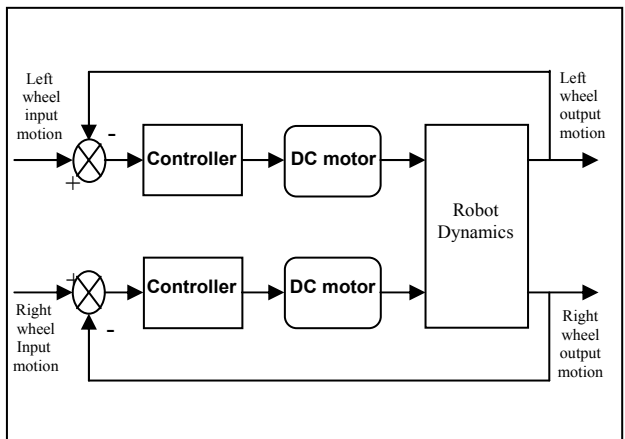

Fig. 5 Block diagram of mobile robot control system

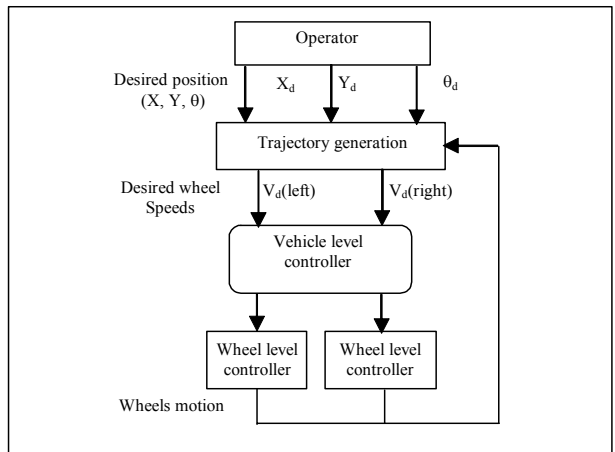

Fig. 6 Motion control of the mobile robot

The self-tuning adaptive control structure offers a good framework to estimate the model parameters. This can be achieved through the estimation mechanism. Then the controller is implemented and is able to cope with the dynamic changes and take the proper action to control the motion. However, it can not estimate or cope with the changes in the surface conditions. The basic block diagram for the self-tuning adaptive motion controller is shown in Fig. 7.

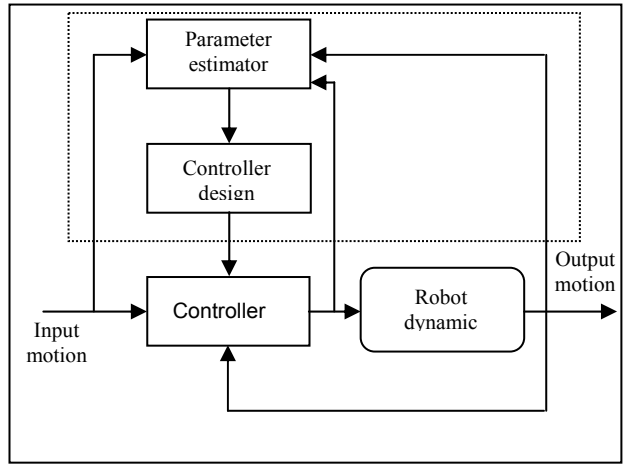

Fig. 7. Self-tuning adaptive controller structure

*4.2. Combined control strategy*
In this strategy, a combined motion control base on 'Low' and 'High' level controls is adopted and presented in this section. The combined controller takes in consideration both the robot dynamics and the environment structure. This is the main fundamental difference between the combined control and the Low level control which does not include the environment structure and its conditions as shown in figure. The two main issues to look for, when including the environment structure, are the presence of obstacles in the surrounding and the condition of surface such as smoothness and dryness. This paper concentrates and emphasizes on the state of the surface in order to examine the robot motion under different surface conditions compare the performance with the Low level control strategy. Fig. 8 shows the robot movement under different surface conditions.

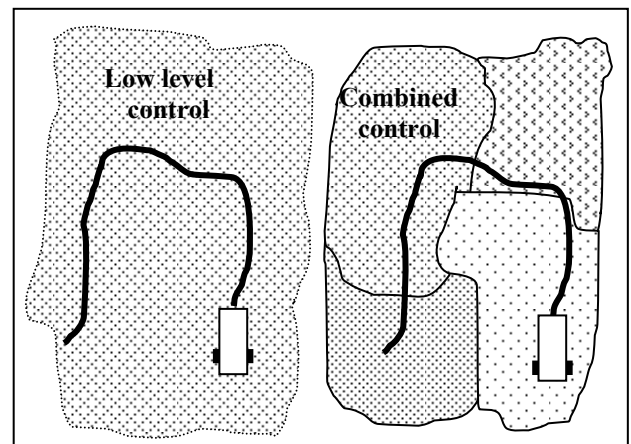

Fig. 8 Robot motion under different surface conditions

## 5. Simulation and Results

A general dynamic model to represent the vehicle as a three dimensional rigid body has been built using Matlab software. The model will calculate the mass, centre of gravity and the inertia of the entire body. These values are used in the dynamic model of the robot which the controller operates on. An arbitrary over all path has been used to test the motion control of the robot. The effect of the 'low' level adaptive and the combined controllers has been investigated on the robot model with changes in the state of the surface condition (i.e. different friction coefficients).

The results indicate that the combined 'low' and 'high' level adaptive controller is able to compensate and cope with the changes on the path condition and provide better path tracking. Selected sections of the velocity trajectories for the right and left wheels have been chosen to illustrate the difference between the performance of the 'low' level adaptive controller and the combined controller under these conditions. Also the displacement error in the over all path-tracking are presented to show the advantages of the combined controller.

Fig. 9 shows the effect of the low level controller. Meanwhile Fig. 10 shows the performance of the combined controller on the vehicle with the following parameters:

mass (m)= 40 [Kg]; Inertia (J)= 2.627 [Kg/m$^2$]; $L_F$= 0.3[m]; $L_R$ = 0.1 [m]. Both controllers subjected to the changes in the surface condition (i.e. friction coefficient). The technical specification for the vehicle motion is subject to some practical constraints which the controller must take into account. These constraints are:

Maximum linear velocity:     $V_{max}$=2 (m/s)
Maximum linear acceleration: $A_{max}$=1.5 (m/s$^2$)
Maximum angular velocity:    $\omega_{max}$=2 (rad/s)
Maximum angular acceleration: $\omega_{max}$=1.5 (rad/s$^2$)

The actuators are sufficiently powerful to achieve these maxima within their own operating constraints on maximum voltage.

Fig. 9(a) shows the path-tracking of the robot. It can be seen that once the slippage occurred, the robot could not follow the desired path. The velocity profile for the chosen section is shown in Fig. 9(b). Fig. 9(c) shows the displacement error of the robot. Here, the error or the deviation of the robot from the desired path is large due to the slippage and the low-level controller could not cope with the changes in the surface conditions. Meanwhile, Fig. 10(a) shows the path-tracking of the robot in the case of the combined controller.

It can be seen that the robot is following the desired path and achieving a good tracking with accuracy. The velocity profile of the chosen section is shown in Fig. 10(b). Fig. 10(c) shows the displacement error which is very small and the robot reaches the final position with accuracy.

## 6. Conclusion

In this paper two control strategies are developed, and tested on the robot. The 'low' level controller performance deteriorated with the changes in the surface condition such as the traction condition (friction coefficient). Meanwhile the combined controller detects the changes and copes with them in an adequate manner, maintaining a largely consistent performance. Some of the issues concerning the environmental structure and the high level control have been presented. Determining the location of the mobile robot plays a vital role in maintaining fast, smooth path-tracking. Measuring the position of the robot in the workspace gives the high level controller an indication of whether the robot is experiencing any slippage or not. All these issues are important for the motion control. The combined control system has been investigated and tested on the differential drive mobile robot. Simulation results show that the performance of the mobile robot under the combined system has improved and the accuracy of path-tracking also improved significantly as it can be seen from the figures.